# Exploring Motivations for Algorithm Mention in the Domain of Natural Language Processing: A Deep Learning Approach

Yuzhuo Wang[1], Yi Xiang[2], Chengzhi Zhang[2]

1. School of Management, Anhui University, Hefei, 230601, China,

2. School of Economics and Management, Nanjing University of Science and Technology, Nanjing, 210094, Jiangsu

**Abstract**: With the formation of the fourth paradigm of scientific research, algorithms have become increasingly important in scientific research. In academic papers, algorithms may be mentioned by scholars with various motivations, using, comparing, or improving algorithms to solve complex research tasks. Identifying these motivations can help scholars discover the relationships between algorithms and further assess their roles and values. Therefore, taking the field of natural language processing (NLP) as an example, this article proposes a complete method to conduct the identification, distribution, and evolution of motivations for mentioning algorithms at the sentence level. Specifically, using manual annotation and machine learning methods, we identify algorithm entities and sentences in the full text of papers, classify motivations for mentioning algorithms by pre-training models and data augmentation techniques, and finally analyze the distribution and evolution of motivations. The results show that the deep learning models trained with the augmented data outperform the traditional machine learning models in the classification task. In academic papers, more than half of the sentences show the direct use of algorithms, while the lowest percentage of motivations are improving algorithms, and the diversity of motivations has been increasing with time. For specific algorithms, grammatical algorithms are mentioned more by the motivation of "description," while more motivations of "use" are found in the machine learning algorithms category. As time passed, the "use" motivations gradually replaced the "description" motivations for different algorithms, and the number of motivation types decreased significantly. Our research explores the identification, distribution, and evolution of authors' motivations for mentioning algorithm entities, which could provide a basis for future algorithm relationship identification and influence evaluation using motivations.

## 1. Introduction

Nowadays, AI (Artificial Intelligence) technology is rapidly changing how humans work, learn, and live at an astonishing pace (Azoulay, 2019), on which algorithms, computing power, data, and knowledge are regarded as the four core elements driving technological advancements (Tang, 2021; ByteBridge, 2021; Deng et al., 2020). Among these four elements in the era of AI, algorithms stand out as particularly important because using computers, anytime and anywhere, relies on algorithms, and the revelation of "indirect and implicit" associations between data and knowledge also depends on algorithms (Hickman, 2013). In scientific research, algorithms are a crucial focus. With the emergence of the fourth scientific paradigm, data-driven research places higher and greater demands on algorithms, which need continuous optimization and innovation to identify, process, and analyze massive amounts of data (Balcan, 2020; Gupta & Roughgarden, 2020). Algorithms also serve as essential research tools, assisting scientists in more efficiently addressing various scientific challenges, whether in the humanities and social sciences or the natural sciences (Abbott, 2017; Kogan et al., 2021).

As a significant scientific product, algorithms have been created, mentioned, used, and discussed in many research papers, and their importance in the knowledge dissemination process is gaining attention from scientists (Zhang et al., 2023). Previous work has explored the evolution path (Zha et al., 2019) and academic influence (Ding et al., 2019) of algorithms using full-text research papers, helping scholars to understand and select algorithms. However, when investigating algorithms in academic texts, scholars mainly focus on syntactic content, such as whether an algorithm is mentioned, its mention location, mention frequency, etc., while neglecting the rich semantic information that academic texts attribute to different algorithms. Specifically, even if algorithms are mentioned with the same frequency in the same section, there can be differences in how authors depict their emotions, functions, and purposes: Some algorithms are introduced as background, while others are directly used to solve the problems, and some algorithms are not only used but also improved as baselines or compared with other algorithms. If we decipher the semantic nuances of texts where algorithms are mentioned, researchers will get a more profound insight into algorithms' interrelationships, prominence, and academic influence. Therefore, we probe the semantic characteristics of algorithms within academic publications by analyzing authors' motivations for mentioning them. Based on these intricate motivations, we can find the disparities in the significance and evolution of various algorithms, which could pave the way for enriched perspectives in forthcoming endeavors related to algorithm evaluations, retrievals, and recommendations.

Taking the field of Natural Language Processing (NLP) as an example, we extract algorithms from academic papers and automatically classify the motivations behind their mentions. Subsequently, we explore the distribution and evolution of motivations within the domain and specific algorithms. The definition of an algorithm evolves, in context, and in the target audience (Blass and Gurevich, 2004; Lum and Chowdhury, 2021). It ranges from considering an algorithm as a program (Cooper, 1969) to a set of rules (Stone, 1971) and a method of data analysis in the AI domain (Genics, 2019). Cormen et al. (2009) provided a more detailed definition of algorithms in the computer science domain: "Informally, an algorithm is any well-deified computational procedure that takes some value, or set of values, as input and produces some value, or set of values, as output.", which inspired our approach to defining algorithms, as we focus on algorithms in the NLP domain. Moreover, considering the close relationship between machine learning algorithms and machine learning models, models that essentially have an input-output process similar to algorithms, or those that include algorithms, are also considered as algorithms. Therefore, this paper argues that in the field of NLP, an algorithm refers to an operational process that incorporates a set of inputs and outputs and that is explicit and efficient, incorporating some of the machine learning models, but distinct from software and application systems. Furthermore, algorithms are mentioned in various forms within academic papers. One form is algorithm metadata, which includes specific descriptions of algorithm procedures or pseudo-code within the paper (See Figure 1). The other form refers to algorithm entities, which are nouns or noun phrases used to denote algorithms. This study focuses on the latter, specifically identifying and discussing the motivations behind mentions of algorithm entities in academic papers.

**4.1 Optimal Histogram Construction**

Consider a sequence of data points indexed by $1 \ldots n$. The data points within each bucket are represented by their mean. We refer to the optimal histogram consisting of $B$ buckets as the optimal $B$-histogram. A basic observation is that if the last bucket contains the data points indexed by $[i+1, \ldots, n]$ in the optimal $B$-histogram, then the rest of the buckets must form an optimal $B-1$-histogram for $[1, \ldots, i]$. It is easy to observe that otherwise the cost of the solution can decrease by taking the optimal $B-1$-histogram and the last bucket defined on points $[i+1, \ldots, n]$. Since the last point, $n$, must belong to some bucket, we search over the $n-1$ bucket boundaries. To put things together, we must solve the problem for $2 \leq k \leq B-1$ for all points, since the bucket boundary for the last bucket can be anywhere between 2 and $n^4$. Given that a bucket is defined by $[i, \ldots, j]$, the error $\text{SQERROR}[i, j]$ in it, is given by

$$\text{SQERROR}[i, j] = \sum_{\ell=i}^{j} v_\ell^2 - \frac{1}{j-i+1}\left(\sum_{\ell=i}^{j} v_\ell\right)^2 \quad (2)$$

Algorithm OptimalHistogram()
*Compute* $\text{SUM}[1, i]$ *and* $\text{SQSUM}[1, i]$ *for all* $1 \leq i \leq n$
*Initialize* $\text{HERROR}[j, 1] = \text{SQSUM}[j, n]$, $1 \leq j \leq n$
*1. For j=1 to n do*
*2. For k=2 to B do*
*3. For i=1 to j-1 do*
*4.* $\text{HERROR}[j, k] =$
$\min(\text{HERROR}[j, k], \text{HERROR}[i, k-1] + \text{SQERROR}[i+1, j])$

**Figure 1. Algorithm metadata in sentence form and pseudo-code form (Guha & Koudas, 2002)**

Using algorithms in the domain of NLP as our focal point, this paper pioneers exploring how specific algorithm entities are mentioned in academic research. Our work has made the following contributions: First, unlike conventional research, which identifies motivation for citing literature, we target a more nuanced understanding of knowledge used patterns in academic research. Based on the deep learning methods, we encompass a holistic research trajectory, from data acquisition to motivation categorization and analysis, using algorithms as examples to provide scholars with a new perspective on dissecting the application of knowledge in academic research. Second, we discern patterns in scholars' algorithm applications by scrutinizing the rationales behind algorithm mentions in specific domains. Such insights pave the way for recognizing shifts in research paradigms and evolving requirements in the NLP domain. Third, by charting the trajectory of algorithms mentioning motivations, from their introduction, use, comparison, and improvement, we can understand the process of their emergence, development, and decline, which equips scholars with tools for a more thorough relational investigation and impact assessment of algorithms in future studies.

## 2. Related work

Using the content of academic papers, we identified and classified the motivations of algorithm mentions. In this section, we introduced two aspects of related work: one focused on exploring algorithms mentioned in academic papers using full-text content, and the other addressed the work related to citation motivations in academic papers.

### 2.1 Explorations of algorithms mentioned in academic papers

Algorithms have become essential subjects and tools in research across various fields. The availability of full-text content and advancements in text mining technology have made it possible to access algorithms mentioned on a large scale, providing resources for further exploration of algorithms' role, value, and impact in academic research. Current research on algorithms mentioned in academic papers includes algorithm recognition, algorithm mention frequencies, locations, and algorithm relationship evolution.

Essential algorithm recognition is carried out using rule-based methods, where scholars can construct rules based on the text features of algorithm metadata. For instance, Bhatia et al. (2011) used regular expressions to identify algorithm pseudo-code titles in papers, confirming the location of pseudo-code and then extracting it. Wang and Zhang (2018) focused on algorithm entities, constructed an abbreviation and alias dictionary for the top-ten text mining algorithm, and extracted these algorithms from academic papers using dictionary matching. After that, they summarized sentence patterns for mentioning data mining algorithms and used these patterns to identify other types of method entities in articles (Wang & Zhang, 2019). However, the ways of describing algorithms in academic papers are not consistent. Tuarob et al. (2013) found that approximately 26% of algorithm pseudo-code does not include a title, so existing rules were unsuitable for handling new mention patterns; they subsequently used a combined approach of rule-based methods and statistical machine learning to extract algorithm pseudo-code. However, the feature engineering required by statistical machine learning methods requires substantial construction costs. Safder et al. (2020) attempted to use deep neural network models to identify metadata sentences that describe algorithms that no longer rely on manually constructed features, and it achieved a 16% increase in accuracy compared to the best statistical model. Integrated methods have also been employed for extracting algorithm entities, Wang et al. (2021) used the integrated statistical machine learning and deep learning model in their research, where Bi-LSTM +CRF significantly improved extraction effectiveness compared to the single model.

After identifying algorithm metadata and algorithm entities, scholars further analyzed different features of algorithms in academic papers. Some studies focus on algorithm sentences; for example, Tuarob et al. (2006) used statistical machine learning methods and neural network models to classify the citation functions expressed in algorithm citation sentences. Other scholars research on individual algorithm entities. Wang et al. (2018) counted the mention frequency of the top ten data mining algorithms and analyzed the depth and breadth of academic influence. Next, they further identified the locations where algorithms are mentioned and found that algorithm entities primarily appeared in the "Method" and "Introduction" sections. Subsequently, they incorporated time into the syntactic features of algorithm mentions and discussed the evolutionary trends of algorithm academic impact (Wang & Zhang, 2020). Beyond individual algorithm entities, scholars further investigated relationships between algorithm entities. Zha et al. (2019) studied hierarchical relationships and comparative relationships between algorithm entities, creating evolutionary graphs for algorithms within the same category.

## 2.2 Motivation analysis based on the content of academic papers

Existing research on authors' motivations in academic papers is concentrated on identifying the motivations of citation content. The primary step is developing category frameworks aimed at classifying these motivations. As early as the 1960s, Garfield (1964) outlined a comprehensive taxonomy of fifteen citation motivations among researchers. Subsequently, Moravcsik and Murugesan (1975) condensed disparate citation motivations into four binary classification categories: conceptual or operational, organic or perfunctory, evolutionary or juxta positional, and confirmative or negational, and then classified motivations based on the citation content. In addition to the citation content, some scholars directly corresponded with the original authors to construct the motivation patterns.

Vinkler (1987), for instance, employed a questionnaire-based approach to investigate the citation behaviors of twenty scholars, thereby classifying citation motivations into professional motivations, connectional motivations, and negative motivations, encompassing both adversarial and apathetic motivations. The above research implicitly assumes that scholars engage in citation practices earnestly and effectively. However, Thorne (1977) posited a contrasting perspective, suggesting that authors may cite irrelevant literature for various reasons. Building upon the foundational research, Teufel et al. (2006) extended the understanding of citation motivations and sentiments by classifying these motivations into four overarching categories and twelve subcategories, further grouping these twelve criteria into three broader categories: negative, neutral, and positive motivations.

According to the classification framework, scholars classify the motivations manually or automatically. Some researchers used interviews or questionnaires to survey the authors and obtain their reasons for citing specific references (Harwood, 2009). Nevertheless, the manual method is challenging to implement, and yields limited data due to its complex processes and labor-intensive nature. Some scholars turned to manual annotation of citation content to identify citation motivations, which streamlines the data acquisition process but still faces constraints regarding data volume, processing speed, and scalability (Moravcsik & Murugesan, 1975). Researchers increasingly explored automated ways for larger-scale data processes to overcome the limitations of manual methods. Garzone and Mercer (2000) were pioneers in using a rule-based approach; they developed 15 lexical matching rules and 14 extraction rules, achieving commendable results at the time. Subsequently, scholars ventured into machine learning, where traditional approaches focused on feature extraction and expansion. For instance, Teufel et al. (2006) extracted various features, including clue words and verb tenses, from citation contexts and combined these features with the IBK algorithm. Abu-Jbara et al. (2013) highlighted the significance of lexical and structural features in the classification process and underscored that support vector machines outperformed other statistical machine learning methods. Jurgens et al. (2018) expanded the features into three categories, and the combination of “pattern-based features, topic-based features, and prototype parameter features” resulted in a macro-$F_1$ score of 54.6% on the ACL-ARC dataset. Recognizing the time-consuming nature of feature engineering, Su et al. (2019) proposed a single-layer convolutional neural network as a more efficient alternative, yielding superior results. Cohan et al. (2019) then integrated 100-dimensional GloVe vectors trained on the Wikipedia corpus as word embeddings, which were combined with ELMO vectors for textual representations. They harnessed the Bi-LSTM model to perform citation motivation analysis on the ACL-ARC dataset, achieving a notable macro $F_1$ score of 67.9%. Based on the identified citation motivations, scholars further evaluated the cited objects, including differences in their contributions, disruption, and the reasons for the differences (Hou et al., 2023; Liu et al., 2023).

Upon reviewing the relevant work, it is evident that scholars have employed various strategies for identifying algorithms and their features mentioned in full-text academic papers. However, current research treated algorithms mentioned within the same article as equivalent entities, disregarding the semantic information within the textual context of algorithms, and diverse intentions underlying scholars’ mention. Regarding the research on authors’ motivation expressed in the paper content, the attention centered on citation motivations. Various methods were employed for motivation categorization, including social surveys, rule-based classification, and machine learning models. The result for automatic motivation classification still hovered at around a 70% $F_1$ score, necessitating substantial training data. Therefore, this paper explored the authors' motivations for mentioning algorithm entities in academic papers, which was called mentioning motivations. We incorporated pre-trained models and data augmentation techniques to enhance classification effectiveness while reducing resource demands. Furthermore, we analyzed the features of motivations mentioning algorithms, aiming to shed light on the roles of algorithms within academic research from a motivational perspective.

## 3. Methodology

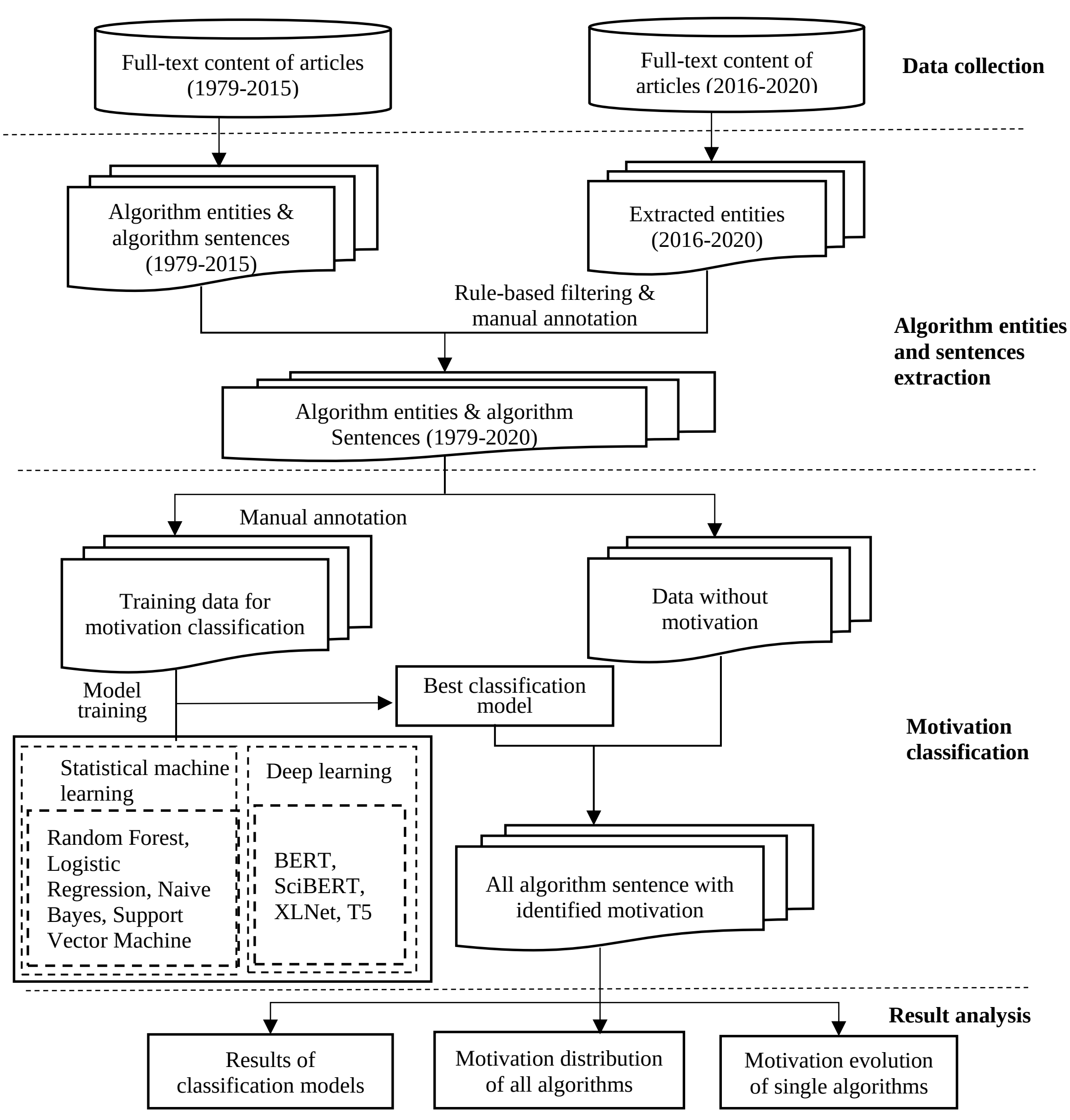

**Figure 2. Flow chart of the motivation classification and analysis**

We first automatically extracted algorithm entities and constructed algorithm dictionaries, then used the dictionary matching method to obtain sentences mentioning algorithm entities from papers. Based on the algorithm sentences, we built a type framework for identifying authors' motivations and annotated motivations of some algorithm sentences as a training corpus. Afterward, we trained various machine learning models to recognize the motivations in the remaining algorithm sentences for the subsequent analysis. The overall flow of our study is shown in Figure 2.

### 3.1 Collecting full-text content of academic papers

We opted to utilize the conference papers from the Association for Computational Linguistics (ACL) as the foundational dataset, because, within the discipline of computer science, there's a prevailing emphasis on disseminating recent findings via academic conferences over traditional journals (Qian et al., 2017). Of the myriad conferences dedicated to NLP, the ACL emerges as a seminal forum. The corpus of papers presented at ACL exemplifies the zenith of contemporary NLP research and encompasses a broad spectrum of sophisticated algorithms. While there are other esteemed NLP conferences, such as EMNLP (Empirical Methods in Natural Language Processing) and NAACL (Annual Conference of the North American Chapter of the Association for Computational Linguistics), the ACL boasts an uninterrupted sequence of events, thereby ensuring a consistent trajectory of research data, making it an ideal source for evolutionary analysis.

We downloaded all data from the ACL Anthology Reference Corpus (https://acl-arc.comp.nus.edu.sg/) and the ACL Anthology Repository (https://github.com/acl-org/acl-anthology), which provided full-text papers from two distinct periods: 1979-2015 and 2016-2020. During our preliminary data processing, we discerned certain anomalies in content and formatting within the XML datasets; therefore, we invited six undergraduate students to isolate these papers, which was followed by rigorous manual scrutiny and content correction by juxtaposing the XML data against the original PDF manuscripts.

### 3.2 Recognizing algorithm entities and algorithm sentences

We trained different models and then used the best model to extract entities that may be algorithms (candidate entities) from the raw corpus. Rule-based filtering and manual evaluation were employed to find the entity words that are explicitly algorithms from the candidate list. We then used lexicon matching to obtain all the sentences mentioning these algorithm entities (algorithm sentences) and used them for the subsequent motivation categorization.

#### (1) Algorithm entities extraction

Wang and Zhang ( 2020) conducted a manual annotation of the ACL conference papers spanning from 1979 to 2015. Their efforts yielded 877 algorithms with 1840 distinct algorithm names and 60,662 algorithm sentences, which were utilized as our foundational training corpus. Specifically, For the separate extraction models, i.e., CRF and Bi-LSTM, we randomly divided the data into training and test sets in the ratio of 20% and 80%, and tested the performance of the models using 10-fold cross-validation. For the other integrated model, we partitioned the data into training, testing, and validation sets at ratios of 70%, 15%, and 15%, respectively. The performance of each model based on training data is tabulated in Table 1, where the performance of a single model is measured by the average of results at each fold. The performance of single models measured by the standard deviation of metrics, and the performance of integrated models on the training and validation sets are documented in Table A, Table B, and Table C in Appendix section, respectively. It can be seen that deep learning models outperformed traditional machine learning models, and integrated models showcased enhanced performance compared to their singular counterparts.

**Table 1. The performance of extraction models**

| Model | Precision | Recall | $F_1$ |
|---|---|---|---|
| CRF | 0.766 | 0.823 | 0.792 |
| Bi-LSTM | 0.842 | 0.845 | 0.842 |
| Bi-LSTM+CRF | 0.872 | 0.901 | 0.886 |
| BERT+CRF | 0.953 | **0.962** | 0.957 |
| BERT+Bi-LSTM+CRF | **0.970** | 0.959 | **0.964** |

Owing to the stellar performance, we selected the BERT+Bi-LSTM+CRF to process the raw corpus, which includes two parts. Part I: 625,445 sentences, excluding the previously labeled algorithm sentences in ACL papers published from 1979 to 2015. We wanted to unearth potential algorithms that might be overlooked during manual annotations. Part II: 427,470 sentences in ACL papers published from 2016 to 2020. Finally, we identified 28,698 and 29,819 candidate entities from the two raw corpora.

Given the inherent challenge of ensuring 100% accuracy in the entity extraction, refining these words for the subsequent analysis is essential. We first removed 38,661 low-frequency words that appeared in only a single sentence. Next, a doctoral student and a master's student annotated a subset comprising 10% (1,779 words) of the remaining 17,778 candidate words to determine the annotation feasibility. They achieved a kappa coefficient of 0.76 in the consistency test, indicating that a single, experienced annotator would be sufficient for the task. Consequently, the doctoral student annotated the remaining terms, identifying 1,835 definitive algorithm entities and 622 uncertain entities. A professor was later consulted to review the ambiguous entities, further refining the list to include 412 algorithms.

In total, we identified 2,247 new algorithm entities. By amalgamating new entities with existing entities in the training corpus, we assembled 4,087 entities (without duplication). These entities brought three issues: First, the authors mentioned an algorithm with a dozen names. For example, the "support vector machine" is also called "SVM," "support vector machines" and so on. Second, an entity may represent different algorithms. For example, "the BP algorithm" can represent either the "backpropagation" algorithm or the "belief propagation" algorithm. Third, an entity may denote algorithms and some other common nouns. For example, "EM" can represent either the EM algorithm or the M entity ($e_1$, $e_1$, …, $e_m$).

For the first problem, we categorized all the entities into full-name type and acronyms type, based on word length and manual review. We used the edit distance algorithm for full-name words to assess the similarity between different names. If the similarity exceeded 95%, we considered them references to the same algorithm. We then manually reviewed these sets and chose the longest word as the standard name for the algorithm. A full-name dictionary containing 1,652 groups with the standard name and alias was compiled.

Abbreviations cause the other two problems. We extracted the original sentence mentioning the acronym and matched the full-name words that appeared in the same sentence. We first determined whether these full-names and acronyms represented the same algorithm by two rules: firstly, the full-name word appeared in brackets after the acronym or vice versa; secondly, the full-name words had the same combination of initial letters as the acronyms, and then we conducted a round of manual review. If the full names were not in the same sentence, we would expand the search range to the three sentences before and after. If the above methods were ineffective, we would resort to a manual search in the original article. An acronym dictionary was constructed in which the standard names corresponding to each acronym were categorized into groups.

After the processing above, the finalized dictionary encompasses 1,652 standard algorithm entities and 3,810 algorithmic entity names.

#### (2) Algorithm sentences matching

Utilizing a dictionary-based approach, we obtained 172,560 sentences mentioning algorithm entities. However, given the prevalence of acronyms, ambiguities arose post-matching. The first kind of ambiguity is that the abbreviation is not an accurate algorithm. For instance, "CNN" in a sentence was the "Cable News Network" rather than the "Convolutional Neural Networks." Second, the abbreviation could represent multiple algorithms. For example, "L-CRF" in a sentence is needed to determine whether it is "Lifelong CRF" or "linear-chain CRF." Ultimately, we obtained 110,345 sentences containing specific algorithm entities that have been disambiguated. We used the abbreviation dictionary mentioned above to clarify which full name a particular abbreviation in the sentence refers to. If the abbreviation and its full name appear in the same sentence, we replace the abbreviation with the full name. If not, we expand the search range or resort to manual processing.

### 3.3 Annotating the training data for motivation classification

Drawing upon the existing classifications for algorithm citation motivations (Tuarob et al., 2019), this study conducted an in-depth examination of algorithm sentences within ACL papers. We consider identifying the motivations for which an algorithm is mentioned from the sentence dimension rather than the article dimension since an algorithm may be mentioned with different motivations in different sentences of an article. Based on the motivation identified from sentences, the final purpose of an algorithm mentioned in an article can be further obtained. We categorized mentioning motivations into two significant types: used and unused, and then subdivided the use into subcategories: use, comparison, and improvement. The specific definitions and examples for each category are presented in Table 2. The granularity of this type of framework strikes a balance, and it enables clear differentiation between the diverse contexts wherein algorithms are mentioned without introducing complexity that might impede annotation or machine recognition. The broader applicability allowed the framework to be used for other methodological entities, such as datasets, software, or evaluation metrics.

**Table 2. Definitions and examples of motivations for mentioning algorithms**

| Type | Definition | Example |
|---|---|---|
| **Description** | Introducing background and principles about the algorithm and other situations where it is impossible to determine whether algorithms are being used. | <algo>loopy belief propagation</algo> is a message passing algorithm for graph inference. |
| **Use** | Using the algorithm in the article without improvement or comparison. | We have used Deb's <algo>NSGA-II</algo>to generate sets of solutions. |
| **Comparison** | Using the algorithm and comparing it with others. | We can also observe that the <algo>adaptive bootstrapping</algo> method and the rap method perform much better than other methods in terms of f-score. |
| **Improvement** | Modifying the algorithm and then using the modified version. | We provide a novel <algo>random walk</algo> model on this graph that incorporates both topical importance of referring sentences as well as frequency and temporal distance of references. |

**Note**: The algorithm entity in the example sentence is enclosed by "<algo></algo>".

An algorithm can be mentioned in an article for multiple different reasons. If we were to annotate the motive within an article as a unit, it would not only be impossible to obtain a unique result but also cause significant interference with machine classification. Therefore, we adopt a sentence-by-sentence annotation strategy, identifying only the motive for mentioning the algorithm in sentences. Subsequently, other researchers can conduct further exploration based on their own needs. They can select the most complex motive as the purpose for why the authors mentioned an algorithm or focus on the diversity of motivations for further analysis. We engaged two annotators and an expert reviewer in annotating the type of motivations. Specifically, one doctoral student and a senior undergraduate student took on the responsibility of annotation, and a professor to provide guidance and address any queries.

**Pre-annotation Phase:** This initial phase consisted of two iterations, during which 200 algorithm sentences were randomly picked for each round and handed over to the annotators for independent annotation. During the annotation process, we observed that for each algorithm mentioned in an article, besides being directly used by the authors, they also appeared in existing related works, meaning the authors would mention how other researchers used these algorithms. For the algorithms mentioned in this context, annotators conducted careful reading and judgment. If a sentence merely describes how others use the algorithm, then that sentence is labeled as "Description." However, if the author indicates their own use of, or improvement upon, the algorithm while describing how others use it, then that sentence is labeled "Use" or "Improvement." The interrater reliability ascertained by Cohen's kappa coefficient achieved 0.80, confirming that the annotators were competent for separate annotations and that the results were dependable.

**Main annotation Phase:** From the amassed 107,554 algorithm sentences, 6,000 sentences were extracted based on the distribution across articles. These sentences were evenly divided between the two annotators and assigned a motivation for mentioning them. Sentences posing ambiguity regarding the motivational type were escalated to the expert reviewer, and the final decisions on such sentences were reached collaboratively.

**Annotation Outcomes:** Among the 6,400 algorithm sentences that emerged from the annotation phases, 79 sentences were discarded due to encoding errors. A unique observation was that specific algorithm sentences mentioned multiple algorithms, but not all these algorithms shared the same motivation. For instance, in the sentence "We applied a modified version of the HITS algorithm and an SVM trained with pseudo-relevant data for article analysis," HITS is noted as a modified version, while SVM is indicated as directly used. Of the 6,321 retained sentences, 60 exhibited diverse motivations, including 37 sentences where a single algorithm had varied motivations and 23 sentences where multiple algorithms had divergent motivations. Given that such sentences constituted less than 1% of the entire corpus, a figure unlikely to impact the outcomes substantially, they were excluded. Ultimately, 6,261 algorithm sentences, each tagged with a specific motivation, were retained. Table 3 provides a breakdown of sentences by their motivation type.

**Table 3. The number of sentences in different categories**

| Type | The number of sentences |
|---|---|
| Description | 1380 |
| Use | 3800 |
| Comparison | 642 |
| Improvement | 439 |
| Total | 6261 |

### 3.4 Classifying the motivation of sentences mentioning algorithms

We employed traditional machine learning (ML) and deep learning (DL) models for motivation classification. Given the minimal information load within sentences, we did not implement data preprocessing. To be specific, four traditional text categorization models were chosen: Random Forest (RF) (Breiman, 2001), Logistic Regression (LR) (Lei et al., 2019), Naive Bayes (NB) (Chen et al., 2009), and Support Vector Machine (SVM) (Cortes & Vapnik, 1995). Drawing upon Zhang et al. (2022) research, given the minimal information load within single sentences, this research utilized pre-trained models for subsequent classification: BERT (Devlin et al., 2019), SciBERT (Beltagy et al., 2019), XLNet (Yang et al., 2019), and T5 (Raffel et al., 2020).

#### (1) Motivation classification based on traditional machine learning methods

During traditional machine learning model training, varied text representation techniques were employed.

ML Experiment I (One-hot text representation): This paper employed the Information Gain (IG) and Chi-Square Test (CHI) for feature selection. Preliminary tests revealed superior CHI performance, prompting its selection. Term Frequency-Inverse Document Frequency (TF-IDF) was utilized for feature weights to vectorize algorithm sentence text, followed by applying diverse traditional models for classification.

We also tried several features mentioned in existing experiments on citation motivation classification. However, these text features didn't significantly benefit model training. Specifically, we utilized features, including the sentiment score of the sentence, total word count, the part of speech, the tense of words, proportion of nouns, proportion of verbs, proportion of pronouns, et al. When all these features were considered, the best-performing model was RF, which achieved a macro-F1 score of 0.4445, while the worst-performing

SVM only achieved a macro-F1 score of 0.2269. Therefore, these results were thus excluded from the paper's discussion.

**(2) Motivation classification based on deep learning methods**

For deep learning models, the research explored training corpus expansion and trained the models by corpus with different sizes.

DL Experiment I (using the original training corpus): We conducted the experiments using the original training corpus. Cohan et al. (2019) found that raw text representation surpasses additional feature utility for citation motivation classification. Pre-training models, including BERT and GPT, are designed to improve the performance of the downstream tasks by learning a generalized text representation through training on large-scale corpora (Qiu et al., 2020). Therefore, this phase directly identified motivations using fine-tuned pre-training models. Four pre-trained language models were utilized in our experiment, including BERT, SciBERT, XLNet, and T5. Taking the BERT model as an example, we used the BERT-base-uncased version provided by the Transformer library. The original text was tokenized by the tokenizer, and a special <cls> token was added at the beginning of each sentence. The token sequences of different sentences were then padded to the same length, which served as the input to the BERT model. After passing through the 12-layer Transformer structure of BERT, we obtained the 768-dimensional semantic vectors corresponding to each token in the input sentence. We then took the semantic vector of the <cls> token from the last hidden state as the vector representation of the sentence. Finally, we employed a fully connected neural network to project the 768-dimensional vector to a 4-dimensional vector representation. Each dimension represents the probability of belonging to different categories of motivations, with the highest probability indicating the inferred category by the model. Next, the cross-entropy loss function was used to calculate the difference between predicted and actual categories, and then update the weights of the entire network through the backpropagation mechanism.

DL Experiment II (training corpus expansion via data augmentation): To ascertain the data size influence on results, we used data augmentation to increase the corpus, which amplifies training data variety without needing new data collection (Feng et al., 2021). In the NLP domain, popular strategies encompass Back Translation (Sennrich et al., 2016), word substitution with language models (Kobayashi, 2018), and so on. The data augmentation process is depicted in Figure 3. Initially, data was divided into training and test sets at a ratio of 80% to 20%, yielding 5008 training samples. Using the NLPAUG library[2], the training data was expanded by replacing the original word with synonyms and inserting new words. As shown in Table 4, the training set comprises 35,034 entries after augmentation, marking a sevenfold increase, which was then utilized to fine-tune the pre-trained models.

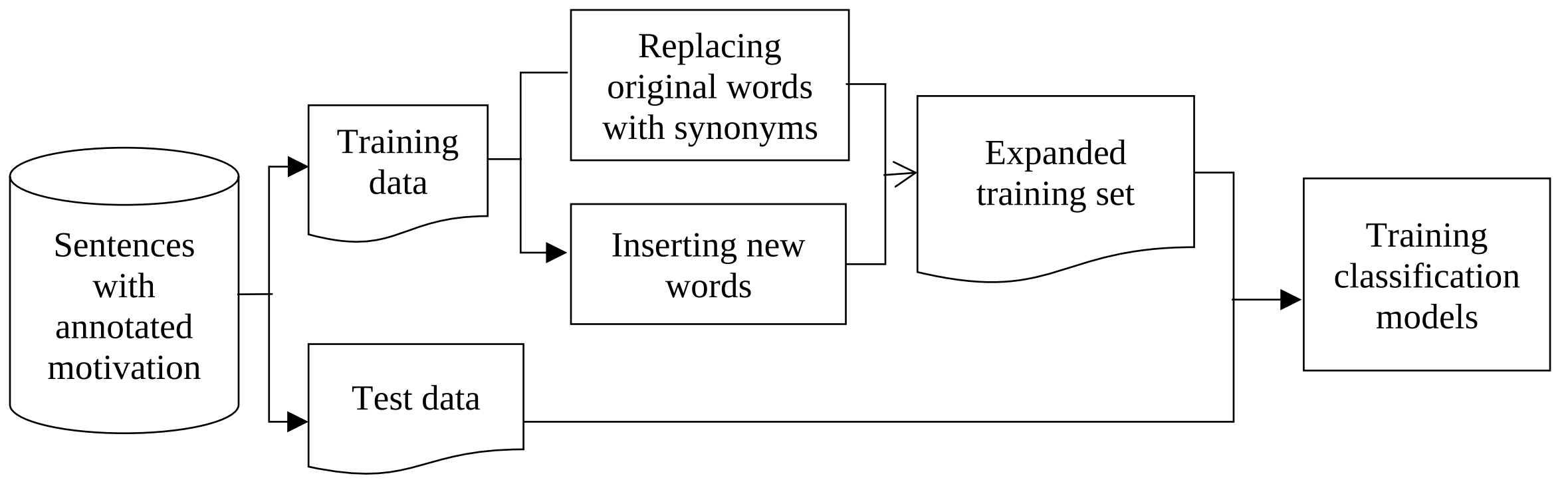

**Figure 3. Data Augmentation of training data for motivation classification**

**Table 4. Examples of sentences after data enhancement treatment**

| Sentence Types | Example of augmented data |
|---|---|
| Original Sentence | Both word embedding models are trained using the Wikipedia corpora. |
| After synonym replacement | Both word embedding models can train with traditional Wikipedia corpora. |
| After inserting new words | Thus, both the word mode embedding models are trained using exactly the relevant Wikipedia corpora. |

## 4. Results

This section details the outcomes of our motivation classification across various models. We then delved into the distribution traits for each kind of motivation. Conclusively, the patterns and progression of motivations mentioning different algorithm entities were investigated.

### 4.1 Results of motivation classification

Leveraging the manually labeled training corpus outlined earlier, we implemented the experimental framework previously described to train traditional ML models and DL models and then assess their classification performance.

#### (1) Performance of traditional machine learning models

Every model undergoes experimentation using a ten-fold cross-validation, whose performance is the average value calculated from the results of each fold and illustrated in Table 5 (the standard deviation of results in each fold is shown in Table D of Appendix). It is evident that when employing the one-hot encoding combined with TF-IDF for text representation, the LR model excels, boasting an accuracy of 0.7178. The performance gap among SVM, LR, and NB is narrow, but RF lags noticeably behind the trio.

**Table 5. Experimental results based on traditional machine learning models**

| Experiment | Model | Macro-P | Macro-R | Macro-$F_1$ |
|---|---|---|---|---|
| ML experiment I: TF-IDF | SVM | 0.7160 | 0.5755 | 0.5966 |
| | LR | 0.7178 | 0.5879 | 0.6096 |
| | NB | 0.7060 | 0.5665 | 0.5892 |
| | RF | 0.6414 | 0.4313 | 0.4608 |

#### (2) Performance of deep learning models

The deep learning models are validated using 10-fold cross-validation, and their performance, which is the average value calculated from the results of each fold, is shown in Table 6 (the standard deviation of results in each fold is shown in Table E of Appendix). The classification of all pre-trained models outperforms the machine learning model. When using only raw data, SciBERT achieves the best results, with a macro-$F_1$ value of 0.7294, which may be because SciBERT is pre-trained from a corpus of scientific literature, and it can represent the algorithm sentence in academic papers with more appropriate features.

After being trained on the more extensive training data, SciBERT is still the best-performing model, which again validates its superiority to academic text data processing. Compared with the previous experiment, the performance of the BERT, SciBERT, and T5 models are further improved, with the T5 model showing a particularly significant improvement. Generally, this section obtained the best-performing dataset and model combination, i.e., the SciBERT model trained on the data-augmented training set, and employed it to classify all the remaining algorithm sentences to obtain the motivation.

**Table 6. Experimental results based on deep learning models**

| Experiment | Model | Macro-P | Macro-R | Macro-$F_1$ |
|---|---|---|---|---|
| DL experiment I: Raw data | BERT | 0.7021 | 0.7275 | 0.7087 |
| | SciBERT | 0.7316 | 0.7332 | 0.7294 |
| | XLNet | 0.7006 | 0.7376 | 0.7119 |
| | T5 | 0.7085 | 0.7335 | 0.7151 |
| DL experiment II: Augmented data | BERT | 0.7114 | 0.7507 | 0.7229 |
| | SciBERT | 0.7398 | 0.7564 | 0.7597 |
| | XLNet | 0.7116 | 0.7257 | 0.7137 |
| | T5 | 0.7339 | 0.7366 | 0.7317 |

### 4.2 Distribution of different motivations in the NLP papers

Using sentences and articles as counting units, we obtained the share of different categories of mentioning motivations in NLP papers. This section also shows the changes in the proportion of different categories and combinations of motivations over time.

#### (1) The number of sentences mentioning algorithms with different motivations

The data culled from the ACL papers highlights some critical trends regarding how algorithms are discussed and utilized within academic texts. From a pool of 107,554 sentences that allude to algorithms, a mere 24% delineate the algorithm itself, with the remainder elaborating on the diverse ways to deploy algorithms. This trend underscores algorithms' instrumental role in NLP research; they are predominantly perceived as mere theoretical constructs and pragmatic tools aiding problem-solving.

For the "used" category, the highest percentage of motivations was "use," followed by "comparison," and the minimum part was "improvement," which is in line with the findings from earlier research on algorithmic metadata (Tuarob et al., 2019). Figure 4 shows the evolution of sentence frequency with different motivations, where "use" is always the motivation with the highest percentage, while "improvement" has the lowest share. After reading some ACL papers, we speculate that the high proportion of "use" may be related to top NLP conference papers focusing on algorithm proposal and usage. Although authors introduce some previously used algorithms in their papers, they more often mention how they utilize existing methods or create new ones and the effectiveness achieved (these sentences are considered evidence of the authors using algorithms in our study). From 1979 to 1985, there was an evident decline and then an increase in the rate of "use," while the other categories of motivation showed the opposite evolution; this apparent fluctuation did not occur after 1985. We think this period corresponds to the formative years of NLP research, where academic conventions around algorithm description might not have crystallized. From 1986 to 2000, the "improvement" references plateaued, but other motivations witnessed minor fluctuation. Beyond the turn of the millennium, the proportions related to algorithm mentions stabilized across all motivation categories, implying that the culture around algorithm discourse in NLP papers had reached a consensus. New algorithms surely burgeoned, but the reasons for their mentions, be it direct application, improvement, comparison, or brief mention, stayed consistent in proportion.

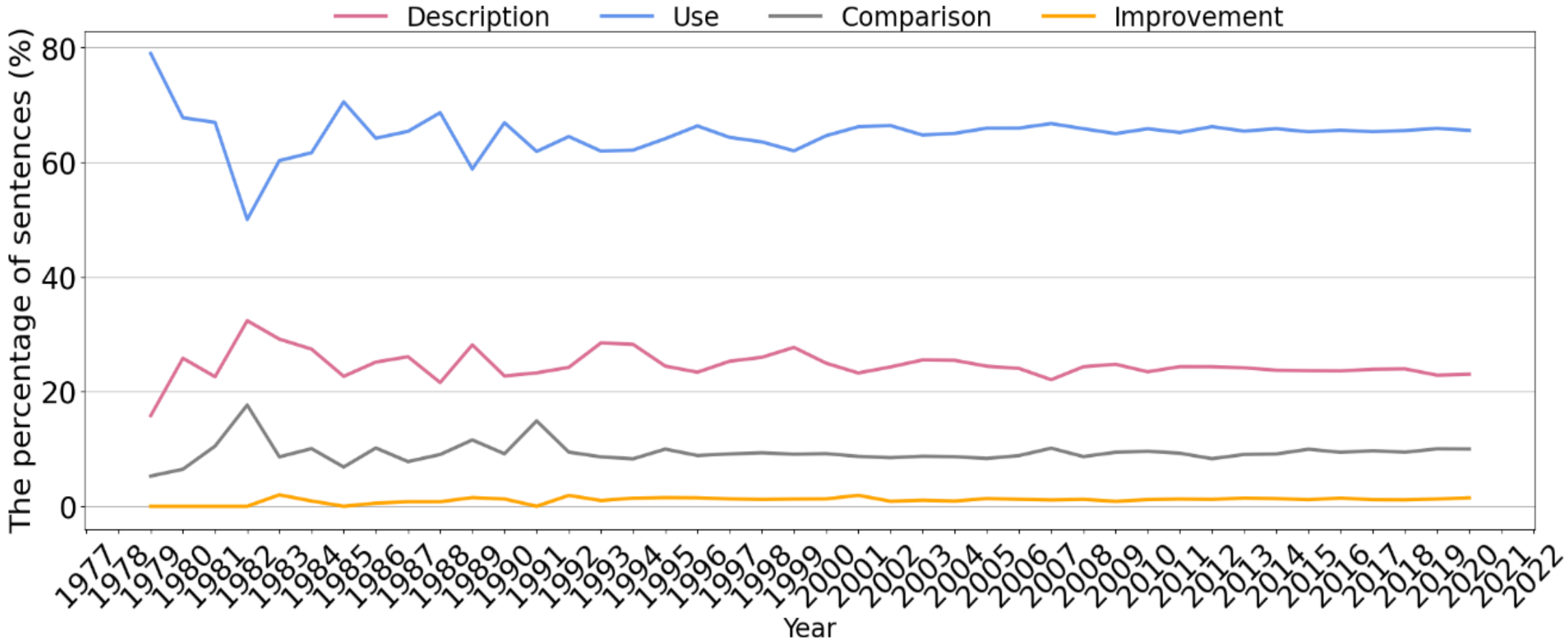

**Figure 4. The percentage of sentences mentioning the algorithm with different motivations each year**

**(2) The combination of motivations of mentioning algorithms within papers**

To investigate the hypothesis from the preceding section regarding the multiple motivations for referencing algorithms in an article, we delved deeper into an article-centric analysis. Instead of counting every mention, we focused on whether an article contained a specific motivation for mentioning algorithms, diving further into this, we examined the combination patterns of different motivations in the articles. Four distinct motivations result in 14 potential combinations, and the percentage of articles having different motivation patterns is presented in Figure 5. Predominantly, articles tend to "briefly mention, directly use, and comparatively assess" different algorithms. The following frequent combination is "description and use" of these algorithms. Interestingly, only 14% of the articles exhibit all four motivations. When narrowed down to articles with a singular reason, motivation "use" stands out, whereas a mere 4% of articles lack this reason.

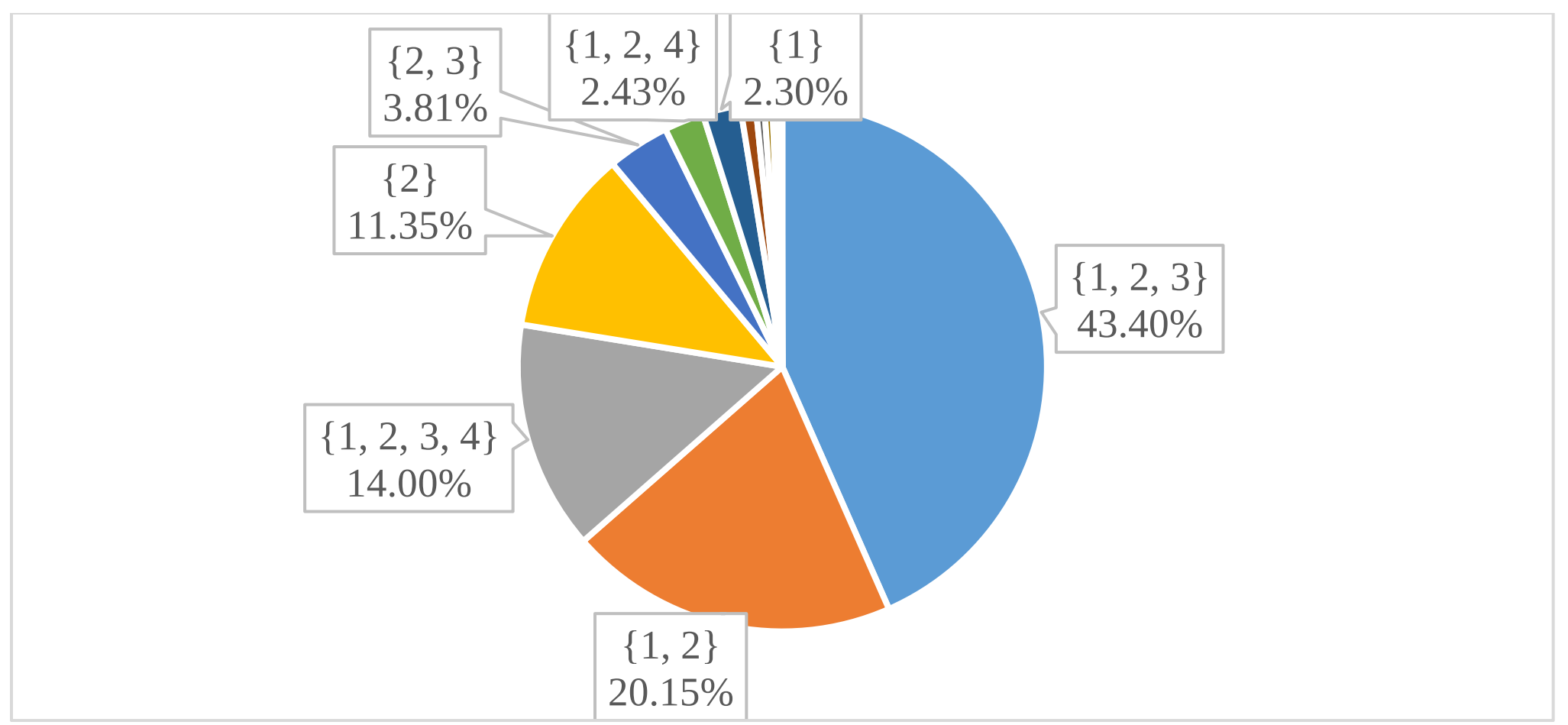

**(Note:1 means the motivation of "description," 2 means the motivation of "use," 3 means the motivation of "comparison," 4 means the motivation of "improvement")**

**Figure 5. The proportion of articles mentioning algorithms in different motivation combination patterns**

We calculated the annual percentage of papers containing different motivations to gain more insights into the evolution of motivation patterns. Note that we only counted one to each motive, no matter how many times it appeared in an article. Figure 6 depicts that the heterogeneity of reasons for mentioning algorithms has undergone a noteworthy transformation. Overall, in different eras, the most common motivations for mentioning algorithms, as well as combinations of motivations, are "comparison," "description, and use," and "description, use, comparison." The proportion of articles with different numbers of motivation types shows varying trends. There's a decline in articles containing just one or two kinds of motivations, particularly articles that mention various algorithms with a single motivation, such as "use" and "comparison," as well as their combination. However, the proportion of articles with three or four kinds of motivations shows an opposite upward trend. The increase in three motivations is primarily due to the growth of "description, use, comparison." We infer that the growth in multiple motivation categories suggests the growing complexity of how algorithms are approached in literature. A deeper temporal analysis indicates that before 2000, there was a stark fluctuation in the number of types, particularly with articles having a single motivation plummeting from 50% to below 20%. The period from 2000 to 2014 saw a modest decline for articles with one or two kinds of motivations, contrasted with an upswing for three or four types. In the post-2014 phase, evolution trends have evolved significantly more drastically. It becomes evident that as the NLP method transitioned from "syntactic class algorithms" to "statistical machine learning" and subsequently to the "deep learning" epoch, there was a broadening in the array of reasons authors had for mentioning algorithms.

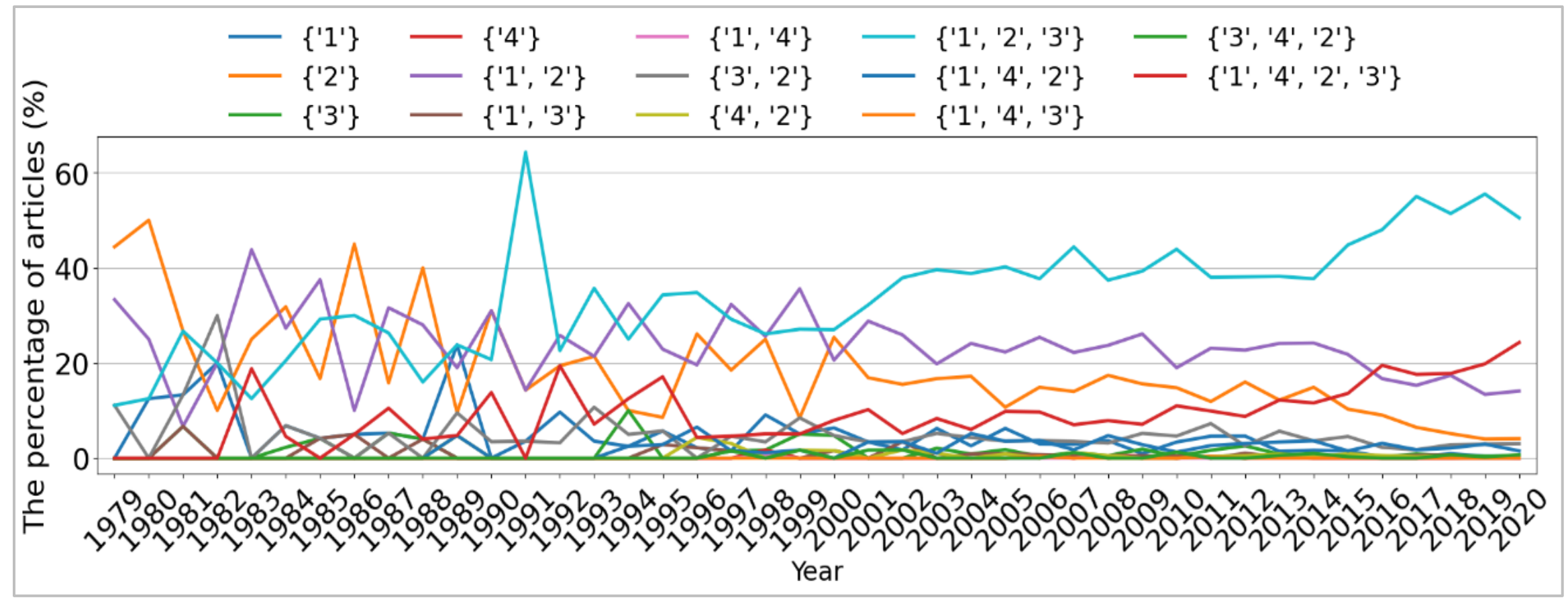

**(Note:1 means the motivation of "description," 2 means the motivation of "use," 3 means the motivation of "comparison," 4 means the motivation of "improvement")**

**Figure 6. The proportion of articles mentioning algorithms with different numbers of motivation types each year**

### 4.3 Mentioning motivations of different types of algorithms

To explore the differences among algorithm classes, we selected a hallmark algorithm from three distinct categories: grammatical algorithms, statistical machine learning algorithms, and deep learning algorithms, to trace the shifts in motivations behind mentioning these algorithms over time. For each stage, based on the evaluation results of different algorithms' influence from existing works (Wang & Zhang, 2020), we selected those with high influence and representative characteristics. Among grammatical algorithms, we chose the Context-Free Grammar (CFG) algorithm, which was powerful enough to express the grammatical patterns of most programming languages (Kadlec, 2008). The Support Vector Machine (SVM) was chosen to represent statistical machine learning algorithms. As described in the "Top 10 algorithms in data mining" (Wu et al., 2008), the SVM has a solid theoretical foundation and is one of the most stable and accurate algorithms among all well-known algorithms. We chose the Recurrent neural network (RNN) from deep learning algorithms. The use of RNNs marked a shift in natural language processing methods towards nonlinear neural network models, which reduce the dependency of sequence tagging models on the Markov assumption and have achieved excellent performance in tasks like machine translation and sequence tagging.

#### (1) The evolution of motivations for mentioning CFG

As for the grammatical algorithms, this research highlighted the Context-Free Grammar (CFG) algorithm, which was proposed in the 1950s. Figure 7 provides a comprehensive overview of its mentions. Figure 7(a) elucidates the percentage of articles referencing CFG based on specific motivations, while Figure 7(b) illuminates the rate of articles containing different combinations of motivations. In the initial decade from 1979 to 1993, CFG witnessed oscillations in its mentions, particularly for "description" and "use." Interestingly, the proportion of articles referencing CFG for singular or multiple motivations was nearly identical. Notably, there is a scant representation of "improvement" linked to CFG, which explains the near-absence of articles embracing all four kinds of motivations.

Changes in motivation experienced a plateau from 1994 to 2013. Articles containing "description" of CFG saw a marked decline, primarily due to the "description and use" decrease. However, the motivation of "use" dominated, accounting for over 80% yearly. After 2013, the share of articles mentioning CFG in a multi-motivation context waned, especially "use and comparison," but singularly-motivation mentions for "description" or "use" grew.

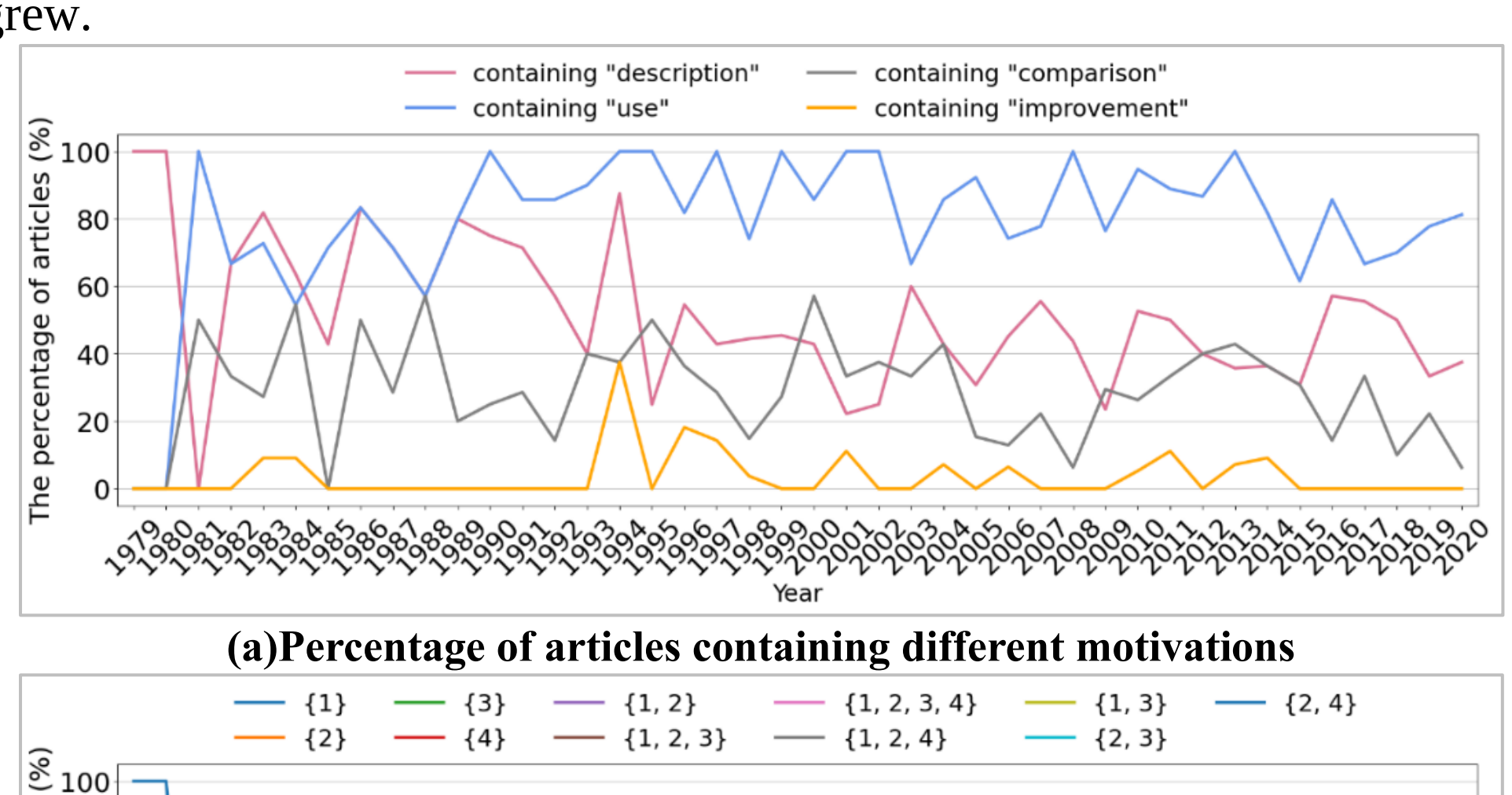

**(a)Percentage of articles containing different motivations**

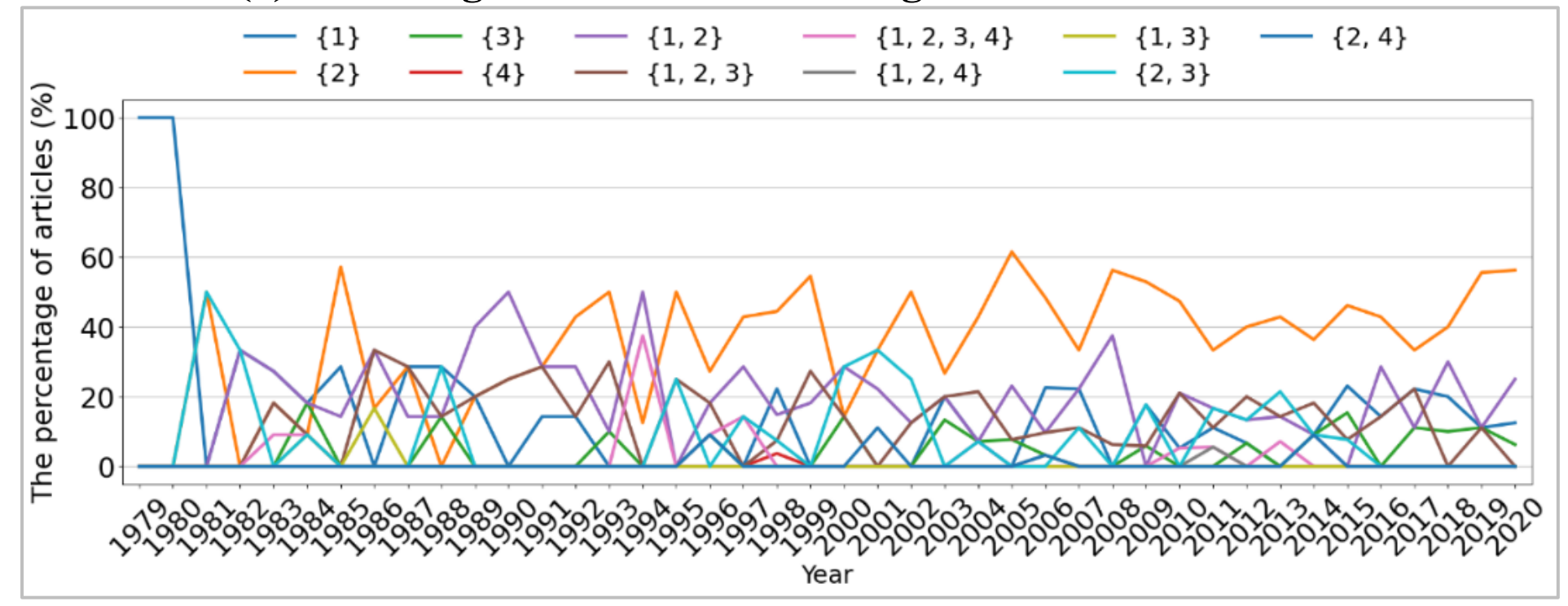

**(Note:1 means the motivation of "description," 2 means the motivation of "use," 3 means the motivation of "comparison," 4 means the motivation of "improvement")**

**(b)Percentage of articles containing different combinations of motivations**

**Figure 7. Distribution of motivations mentioning CFG in different years**

### (2) The evolution of motivation mentioning SVM

We chose the support vector machine as the representative statistical machine learning algorithm. The SVM was first proposed in 1965 and underwent significant improvements in 1995, greatly enhancing its influence. Figure 8 presents the evolution of motivation for mentioning SVM in academic articles. We designated the period from 2000 to 2006 as the developmental phase, when the proportion of articles discussing SVM with simple "description," direct "use," and "comparison" motivations experienced significant fluctuations. There was a noteworthy decline in the prevalence of articles with a single type of motivation, but the share of three kinds of motivations, mainly "description, use, comparison," increased.

Between 2007 and 2014, the "description" category, including "description and use" and "description, use and comparison," experienced a significant reduction. Simultaneously, articles discussing algorithms with "one kind" of motivation increased, especially "use" and "compare," while the proportion of two or three kinds of motivations decreased. We hypothesize that as various types of deep learning algorithms emerged, SVM primarily served as a foundational benchmark for comparison. Following 2015, articles merely describing the algorithm became more common, but those focused on using and comparing SVM algorithms entered a decline phase, while the proportion of articles aiming to improve SVM remained essentially unchanged.

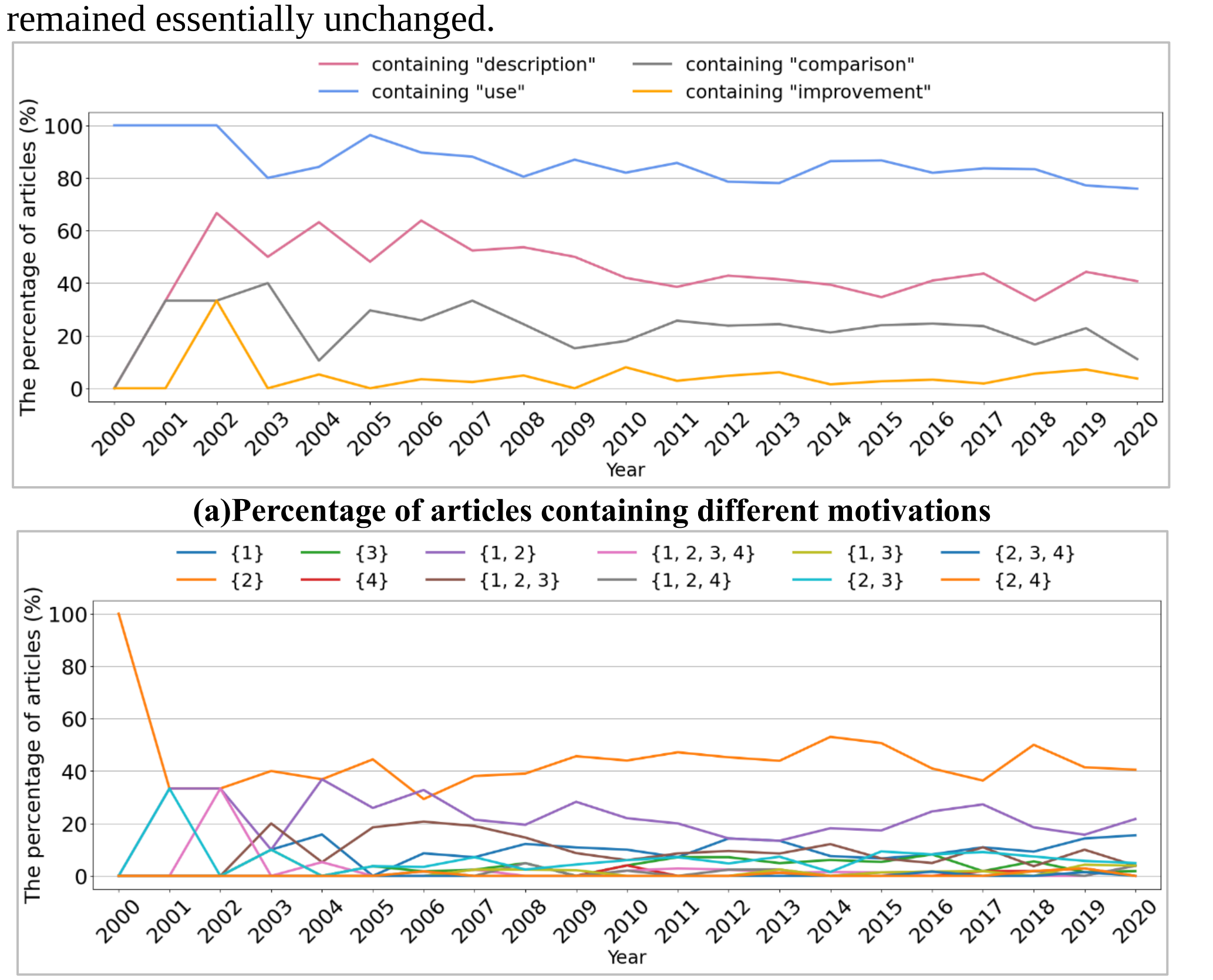
containing "description"
containing "comparison"
containing "use"
containing "improvement"
The percentage of articles (%)
Year
(a)Percentage of articles containing different motivations
{1}
{3}
{1, 2}
{1, 2, 3, 4}
{1, 3}
{2, 3, 4}
{2}
{4}
{1, 2, 3}
{1, 2, 4}
{2, 3}
{2, 4}
The percentage of articles (%)
Year

**(Note:1 means the motivation of "description," 2 means the motivation of "use," 3 means the motivation of "comparison," 4 means the motivation of "improvement")**

**(b)Percentage of articles containing different numbers of motivation types**

**Figure 8. Distribution of motivations mentioning SVM in different years**

### (3) The evolution of motivation mentioning RNN

We chose the Recurrent Neural Network (RNN) as a representative deep learning algorithm. The RNN concept was first proposed in 1986, but it was not until 2006, when a method for rapidly training deep neural networks was introduced, that RNN began to be gradually utilized. As depicted in Figure 9, the initial stage of RNN's development occurred between 2012 and 2015, when the volatility in the proportions of each kind of motivation was more pronounced than in subsequent timeframes.

Between 2015 and 2018, the percentage of articles directly utilizing RNNs remained relatively stable, signifying a developmental phase. However, the share of "description" and "use" experienced a notable decline after 2018; during this period, the proportion of articles containing only two single motivations increased, while those including "description and use" decreased. Regarding the shift in the number of motivational types, once again using 2015 as a pivotal point, we observed a decline in single-type motivation and a rise in multi-type motivation during the earlier phase. Conversely, the opposite evolutionary trend emerged after 2015.

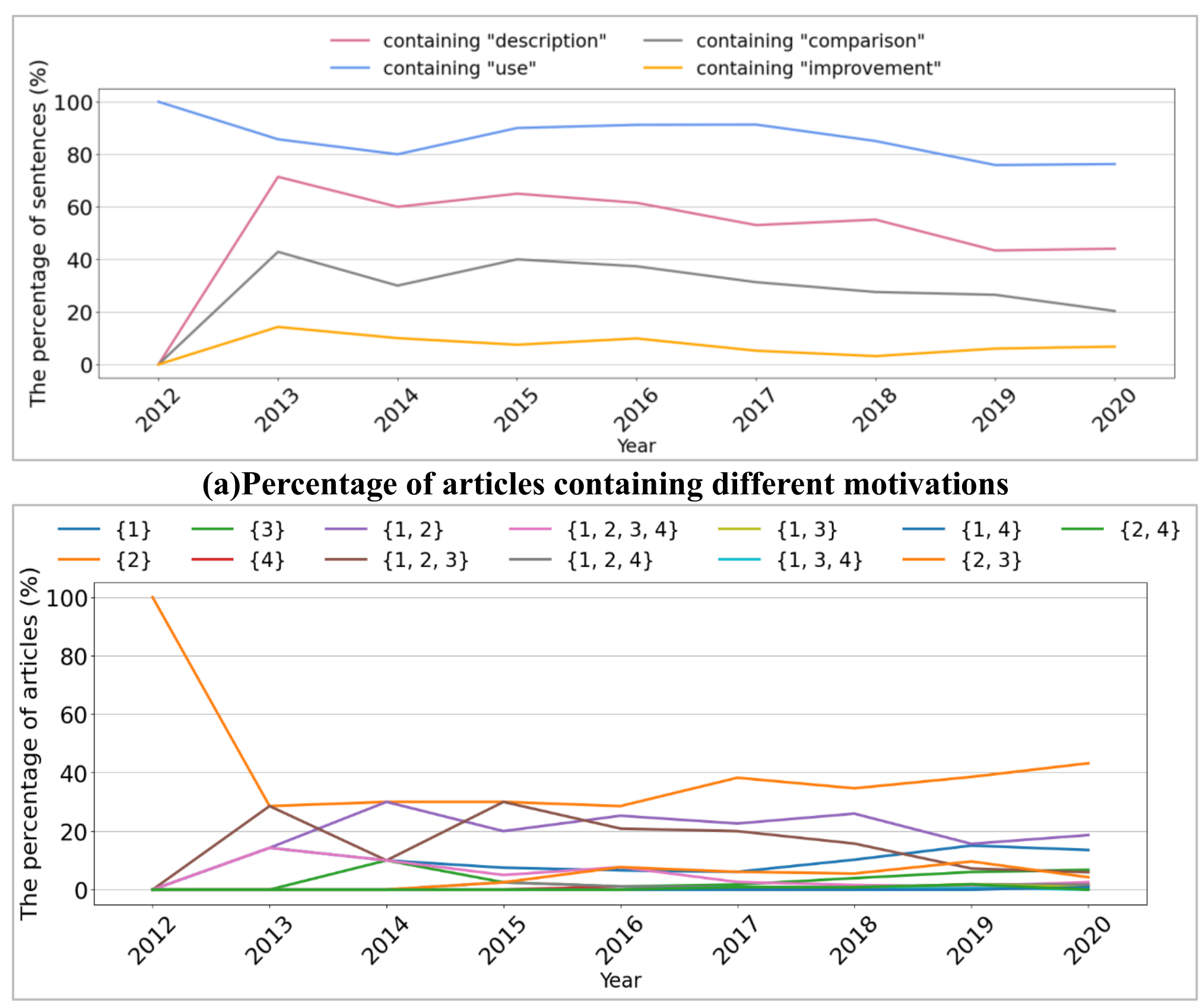

**(a)Percentage of articles containing different motivations**

**(b)Percentage of articles containing different numbers of motivation types**

**Figure 9. Distribution of motivations mentioning RNN in different years**

# 5. Discussion

## 5.1Analyzing the causes of classification errors

In this paper, the most effective motivation classification model is SciBERT. Although the best model achieved the macro-$F_1$ classification score of 0.7528. When compared to prior work on the classification of algorithm metadata motivation (Tuarob et al., 2019), which focused on citation sentences from Citeseer and applied the statistical machine learning models to classify the authors' motivations for citing the algorithms. The results showed that the SVM classifier trained with combined content and context features was the best model, and it achieved a weighted average F1-score of 0.748. Our results demonstrate the improvement, but there is still room for further enhancement. We thoroughly examined prediction errors during both the machine training and testing phases, and then identified some examples to understand the causes of prediction inaccuracies. Table 7 details ten examples attributed to three critical kinds of errors.

### (1) Misinterpretation of core verb

Some algorithm sentences exhibit complex semantics. In cases where the machine fails to accurately comprehend the algorithm entity within the sentence and connect it with preceding verbs, it can lead to misclassification of motivations—for example, sentence eight in Table 7.

Example: *In this paper, we present an extension by using the random walk model to alleviate data sparsity for sp.*

In this sentence, the authors employed the random walk model for an extension, rather than extending the random walk algorithm itself. Therefore, the correct motivation should be "use." However, the machine did not accurately identify the core element of the sentence, namely, "using the random walk model," but thought the motivation was "improvement" based on the "extension" in the sentence.

**Table 7. Examples of motivation prediction errors**

| No. | Correct category | Predicted category | Sentence |
|---|---|---|---|
| **1** | comparison | use | We implemented logistic regression using l-normalization, finding this to outperform normalized and non-normalized versions. |
| **2** | comparison | description | We also intend to compare our approach with other machine learning algorithms using all the kss employed in aleph |
| **3** | comparison | improvement | First, we extend to sentiment classification the recently proposed structural correspondence learning (scl) algorithm, reducing the relative error due to adaptation between domains by an average of 3.7% over the original scl algorithm and 2.45% over a supervised baseline. |
| **4** | improvement | description | Then a modified iterative propagation is carried out over the entire graph to select the most relevant triples of background knowledge to the given source document. |
| **5** | improvement | use | In this paper, we present a novel generative bayesian model that learns domain/dialog act/slot semantic components as latent aspects of text utterances. |
| **6** | description | comparison | There are two major differences between blasso and fslr. |
| **7** | description | use | This makes svm well fitted to treat classification problems involving relatively large feature spaces such as ours. |
| **8** | use | improvement | In this paper, we present an extension by using the random walk model to alleviate data sparsity for sp. |
| **9** | use | description | We first review the path ranking algorithm (pra) as introduced by (lao and cohen, b), paying special attention to its random walk feature estimation and selection components. |
| **10** | use | comparison | We build our classification models using the support vector machine (svm) implementation provided by yamcha. |

### (2) Difficulty in grasping complex comparatives

When predicting sentences whose true motivation is "comparison," the machine often misclassifies them into "use" due to its limited ability to understand more nuanced comparisons, which are exemplified by the first and third sentences in Table 7.

Example: *We implemented logistic regression using L-normalization, finding this to outperform normalized and non-normalized versions.*

Example: *First, we extend to sentiment classification the recently proposed structural correspondence learning (scl) algorithm, reducing the relative error due to adaptation between domains by an average of 3.7% over the original scl algorithm and 2.45% over a supervised baseline.*

In the first sentence, the author highlighted that logistic regression, when normalized, "outperforms" other methods. Here, "outperforms" is the pivotal term indicating a comparison, yet the machine may struggle to discern its comparative connotation accurately. Similarly, in the second sentence, the author employed the phrase "reducing... by" to convey that the "structural correspondence learning algorithm" could reduce error rates and perform better than other models. The machine might not fully grasp the comparative aspect and erroneously interpret it as an indication of improvement.

### (3) Struggles with complex improvement terminology

For sentences with "improvement" motivation, the machine often grasps the usage motivation but falters in interpreting the improvement. Upon reviewing the content of misclassified sentences, we posit that this issue arises because the model can comprehend improvement motivations conveyed through verbs, such as "we modified XXXX algorithm," but encounters challenges when improvement is expressed through adjectives, as seen in the fifth sentence in Table 7.

Example: *In this paper, we present a novel generative Bayesian model that learns.*

In this sentence, the author used the term "novel" to signify an enhancement to the traditional "generative Bayesian model." However, it appears that the machine did not recognize the intent for improvement conveyed by the "novel" and instead categorized it as direct "use."

## 5.2 Potential applications for motivations of mentioning algorithms in academic papers

This study studies the motivations for mentioning algorithm entities in academic papers. Interestingly, the most prevalent motive is "use," as observed in our randomly labeled training corpus and the final classification outcomes. In contrast, research on the citation motivation of algorithm metadata predominantly points to a simple "description" of algorithms as the dominant motive (Tuarob et al., 2019). This divergence implies that algorithm metadata often presents algorithms as foundational or background knowledge, while algorithm entities emphasize the algorithm's practical application as a method or tool. Hence, if one aims to assess the value of algorithms from the perspective of academic productivity, it may be pertinent to extract algorithm entities and pinpoint sentences that refer specifically to these entities.

Our findings underscore the authors' varied objectives when discussing algorithms within a single paper, highlighting the distinct roles that different algorithms play. Therefore, when gauging the academic influence of algorithms, it might be worthwhile to construct factors based on the motivations to conduct a more nuanced measure. Furthermore, discerning the reasons for referencing algorithms can offer insights into their relationships. For instance, algorithms discussed in a comparison context may be rivals, whereas those in the improvement sentences might be collaborators. Additionally, our study reveals temporal shifts in the motivations for discussing algorithms. Over the years, there's been a decline in papers referencing an algorithm for multiple reasons, while a singular purpose initially waned but later saw an uptick. There is also a noticeable dip in papers that mention algorithms with the "simple description" motive. Hence, in the future, tracking the evolving motivations over time may be a potential avenue to classify an algorithm's life cycle and even forecast its future trajectory.

### 5.3 Effect of time on motivation to mention algorithms

The analysis in the preceding section highlights how the publication dates of papers that discuss algorithms can influence the distribution of motivations for mentioning these algorithms. When examining the entire field, the variation in the proportion of sentences with different motivational mentions over the years appears minimal. However, a deeper look at individual algorithms reveals significant shifts in both the types and proportions of motivational mentions. We hypothesize that the relatively minor changes observed in the overall distribution may stem from the presence of classical, developing, and emerging algorithms in different years. These categories represent different stages in the lifecycle of algorithms. As such, the amalgamation of various motivations across these lifecycle stages tends to stabilize the overall distribution of motivations, converging towards a relatively constant percentage. This pattern suggests that while the field as a whole may exhibit stability in motivational distribution due to the blending of different algorithmic stages, the motivations for mentioning specific algorithms are indeed subject to change over time. This dynamic underscores the importance of considering the lifecycle stage of an algorithm when analyzing the evolution of its mentions and motivations in scholarly communications.

For a single algorithm, the motivation to mention it shows different trends when the papers mentioning it are at different stages of its lifecycle. Firstly, regarding the diversity of motivations, unlike the broader domain of NLP, where the diversity of motivations increases over time, the proportion of articles mentioning algorithms with a single motivation decreases first and then increases. In contrast, the proportion of two or more kinds of motivations shows an opposite trend. Secondly, the “use” motivation remains relatively stable when considering specific motivation categories. Conversely, the percentages associated with “description,” “comparison,” and “improvement” exhibit decreasing trends over time, and “description” experiences the most significant decline. At the same time, different classes of algorithms exhibit different distributions of motivation in similar life cycle stages. During the stabilization period of grammar algorithms, the proportions of “description” and “comparison” motivations show little difference, and “improvement” gradually faded after 2014. In contrast, statistical machine learning algorithms and deep learning algorithms exhibit a higher proportion of “description” motivations than “comparison” motivation, and deep learning algorithms display a more pronounced presence of “comparison” and “improvement” motivations.

In summary, Ignoring the nuanced motivations behind the mentions of individual algorithms and focusing only on the aggregate within the domain can lead to misleading conclusions. The above analysis reveals that the distribution of motivations for mentioning algorithms across the field may be influenced by the algorithms' characteristics, their lifecycle stages, and the varying research paradigms of different generations. As the time factor significantly impacts the motivations associated with each algorithm, when developing methods for automatically identifying motivations, incorporating the publication time of the papers as a feature in the model training could potentially influence the model's performance. Therefore, our future studies will further consider these dynamics to avoid confounding the actual influences shaping how algorithms are discussed in scholarly literature.

## 6. Conclusions

This study explores the authors' motivation for mentioning algorithms in academic papers, focusing specifically on the field of NLP. After extensive data extraction of algorithm entities and related sentences, we classify the motivations into four categories: "description," "use," "comparison," and "improvement." Our findings indicate that pre-trained models outperform other models in the classification task. Data augmentation further enhances the model's performance while minimizing the need for manual annotations. Furthermore, our analysis of the distribution of motivations across eras and specific algorithms underscores the pivotal role algorithms play in scientific research. The expectations for algorithm utility in NLP research are becoming increasingly diverse. Moreover, when we inspect the motivations of individual algorithms, we notice that grammatical and classical machine learning algorithms predominantly feature simple "descriptions." In contrast, algorithms garnering a broader spectrum of motivations are typically deep learning methods.

However, as an initial exploration, our study still has some limitations. Firstly, the categorization framework presented is relatively basic. Some nuanced semantic expressions, such as "criticism," have been bucketed under "description" without further differentiation. Future endeavors could aim for a more intricate categorization framework. Secondly, in the automatic identification of algorithm entities and motivations of mentions in papers, the current study randomly assigns papers published in all eras for model training without explicitly considering the effect of the publication time of the papers, leading to possible temporal bias in the results. In our subsequent work, we will try to incorporate time as a feature or influence in the training experiments of automatic identification models. Thirdly, we allocated a single motive to every sentence that mentioned algorithms. Although these singular motivation assignments cover a vast majority, there are instances of extended sentences containing multiple mention motivations. In the future, we might explore strategies to segregate motivations within such complex sentences by breaking down the sentences or considering the relative positioning of algorithm entities.

## Acknowledgments

This paper was supported by the National Natural Science Foundation of China (Grant No.72074113).

## Appendix

**Table A. The performance of single models based on the standard deviation of metrics**

| Model | Macro-P | Macro-R | Macro- $F_1$ |
|---|---|---|---|
| CRF | 0.0976 | 0.1341 | 0.0861 |
| Bi-LSTM | 0.1345 | 0.1144 | 0.1140 |

**Table B. The performance of extraction models based on the training set**

| Model | Precision | Recall | $F_1$ |
|---|---|---|---|
| Bi-LSTM+CRF | 0.9311 | 0.9254 | 0.9224 |
| BERT+CRF | 0.9722 | **0.9910** | 0.9815 |
| BERT+Bi-LSTM+CRF | **0.9746** | 0.9893 | **0.9819** |

**Table C. The performance of extraction models based on the validation set**

| Model | Precision | Recall | $F_1$ |
|---|---|---|---|
| Bi-LSTM+CRF | 0.8938 | 0.9076 | 0.8894 |
| BERT+CRF | 0.9498 | **0.9702** | 0.9599 |
| BERT+Bi-LSTM+CRF | **0.9702** | 0.9588 | **0.9595** |

**Table D. The performance of statistical machine learning models based on the standard deviation of metrics**

| Experiment | Model | Macro-P | Macro-R | Macro-$F_1$ |
|---|---|---|---|---|
| ML experiment I: TF-IDF | SVM | 0.0389 | 0.0426 | 0.0522 |
| | LR | 0.0485 | 0.0565 | 0.0678 |
| | NB | 0.0305 | 0.0404 | 0.0423 |
| | RF | 0.0542 | 0.0445 | 0.0410 |

**Table E. The performance of deep learning models based on the standard deviation of metrics**

| Experiment | Model | Macro-P | Macro-R | Macro-$F_1$ |
|---|---|---|---|---|
| DL experiment I: Raw data | BERT | 0.0191 | 0.0221 | 0.0187 |
| | SciBERT | 0.0341 | 0.0235 | 0.0211 |
| | XLNet | 0.0266 | 0.0213 | 0.0208 |
| | T5 | 0.0272 | 0.0266 | 0.0193 |
| DL experiment II: Augmented data | BERT | 0.0247 | 0.0244 | 0.0136 |
| | SciBERT | 0.0278 | 0.0212 | 0.0162 |
| | XLNet | 0.0330 | 0.0259 | 0.0216 |
| | T5 | 0.0235 | 0.0281 | 0.0201 |